\newif\ifCONF
\CONFfalse

\newif\ifSUBM
\SUBMfalse

\newif\ifDRAFT
\DRAFTtrue

\newif\ifINCEMBED
\INCEMBEDfalse

\ifCONF
\documentclass[twoside,leqno,twocolumn]{article}  has 
\usepackage{ltexpprt}

\else
\documentclass[11pt,letterpaper]{article}
\usepackage{fullpage}
\usepackage{natbib}
\usepackage{makecell}
\usepackage{wrapfig}
\usepackage{caption}
\usepackage{subcaption}

\usepackage[sc,osf]{mathpazo}   
\usepackage[euler-digits,small]{eulervm}
\fi

\ifSUBM

\DRAFTfalse
\fi

\usepackage[obeyFinal, textsize=tiny]{todonotes} 

\usepackage{url}
\usepackage{amsmath}
\usepackage{mathtools}
\usepackage{amssymb}
\usepackage{graphicx}
\usepackage{stmaryrd}
\usepackage{algorithm}
\usepackage{algorithmic}[section]
\usepackage{arydshln}
\usepackage{multirow}

\usepackage[colorlinks,urlcolor=blue,citecolor=blue,linkcolor=blue]{hyperref}

\newcommand{\Reals}{{\mathbb R}}
\newcommand{\mvec}{\mathrm{vec}}
\newcommand{\sketch}{\mathrm{sk}}
\newcommand{\upset}[1]{{\uparrow}\{#1\}}
\newcommand{\OS}[1]{\mathtt{\small OrderSketch}(#1)}

\newcommand{\defeq}{\stackrel{\textit{\tiny{def}}}{=}}

\ifCONF
\newtheorem{definition}{Definition}[section]
\newtheorem{assumption}{Assumption}[section]
\else

\newtheorem{theorem}{Theorem}

\newtheorem{remk}[theorem]{Remark}
\newtheorem{exmp}[theorem]{Example}

\fi 

\def\FullBox{\hbox{\vrule width 8pt height 8pt depth 0pt}}

\def\qed{\ifmmode\qquad\FullBox\else{\unskip\nobreak\hfil
\penalty50\hskip1em\null\nobreak\hfil\FullBox
\parfillskip=0pt\finalhyphendemerits=0\endgraf}\fi}

\def\qedsketch{\ifmmode\Box\else{\unskip\nobreak\hfil
\penalty50\hskip1em\null\nobreak\hfil$\Box$
\parfillskip=0pt\finalhyphendemerits=0\endgraf}\fi}

\ifCONF\else

\fi

\begin{document}
\pagenumbering{gobble}


\title{Order Embeddings from Merged Ontologies using Sketching}
\author{
Kenneth L. Clarkson
\\
IBM Research\\
klclarks@us.ibm.com
\and 
Sanjana Sahayaraj
\\
IBM Research\\
sanjana.sahayaraj2@ibm.com
}

\date{}

\maketitle
\begin{abstract}
We give a simple, low resource method to produce order embeddings from ontologies. Such embeddings map words to 
vectors so that order relations on the words, such as hypernymy/hyponymy, are represented in a direct
way. Our method uses sketching techniques, in particular countsketch, for dimensionality reduction. We also
study methods to merge ontologies, in particular those in medical domains, so that order relations are preserved.
We give computational results for medical ontologies and for wordnet, showing that our merging techniques
are effective and our embedding yields an accurate representation in both generic and specialised domains.
\end{abstract}

\section{Introduction}

While the NLP literature has recently seen one groundbreaking technique after another in feature representation and embedding, there is still work that to be done in imparting knowledge into embeddings,
beyond the use of contextual information; moreover, most of the techniques require huge
computing time and power, while training from scratch. The medical field is an area where feature representations that are interpretable, and that need limited computing resources to design,
can benefit the adoption and acceptance of these systems in practice. Through the $\mathtt{OrderSketch}$ algorithm presented in this paper, we aim to provide a step towards a knowledge-rich feature representation
technique that is based directly on ontologies and is also computation-resource friendly.
The paper is organized as follows: First we present related work on other knowledge representation and order detection techniques. Next we introduce our algorithm $\mathtt{OrderSketch}$, which is simple and yet also effective in capturing
knowledge and order information from ontologies. We apply our embedding algorithm to WordNet, and to
an augmented medical ontology called SnoMeSHNet that we introduce. After presenting our current constructions
and experiments, we discuss future experiments and goals for this work, in progress. 

\section{Related Work}
There has been previous work in representing hierarchical information found in data sources.
\citep{termtaxonomy} predicts hypernym term given a hyponym and context through concatenation of one hot
vector embedding of the hypernym, hyponym, offset vectors of context words and dynamic weighting of a neural
network based on context. There has been work on using a neural network and scoring mechanism to detect
hypernymy relations and differentiate from other relations through distributional similarity in
\citep{hyperdetect}. \citep{unsupdistvec} presents an approach to a reversed form of order embedding,
where an unsupervised embedding method is used to learn embeddings such that the embedding of a hypernym is
larger than that of it's hyponym in every dimension. There have also been approaches such as, \citep{lear},
which is a post-processing mechanism to transform vector space and align vectors to reflect hypernymy
presence and directionality, where ontology knowledge representations like ours can be utilized. 

When it comes to specialty domains such as medicine, \citep{icdHierarchy} shows that hierarchical representation of ICD codes has more information. \citep{HyperbolicEmbed} shows how hierarchical information in ICD9 codes can be incorporated in hyperbolic embeddings and used to reconstruct input to effectively process EHR data.

\subsection{Order embeddings}

In the settings of ontologies, especially,
it may be that the most effective embedding does not represent synonomy,
but other relations between words, such as lexical entailment.
\citep{vendrov2015order} produce maps $g: \mathtt{Tokens} \rightarrow \Reals^n$
such that for tokens $X$ and $Y$, $Y \mathrm{\ entails\ } X \iff g(X) < g(Y)$,
where the inequality means coordinatewise dominance, that is, every coordinate of $g(X)$ is at most
the corresponding coordinate of $g(Y)$. This can be called a conical embedding: if any point
in $\Reals^n$ is a potential concept, the cone
\[  
c(X) \defeq \{ y \in \Reals^n \mid g(X) < y \}
\]
corresponds to the set of concepts that entail $X$. This can also be expressed in
terms of subsets:
\[
Y \mathrm{\ entails\ } Z \iff c(Y) \subset c(X).
\]
Such embeddings extend beyond lexical entailment; for example,
\cite{vendrov2015order} map captions and images to $\Reals^n$,
such that a caption is appropriate for an image if its embedding is
coordinatewise dominated by the embedding of the image. Such an embedding could
also capture entailment among captions: if caption $X$ is appropriate to every
image caption $Y$ is, then the embedding of $X$ should be coordinatewise dominated by that of~$Y$.

Lexical and caption entailment are partial orders that are
transitive: if $X\,\texttt{is-a}\, Y$ and $Y\,\texttt{is-a}\,Z$,
then $X\,\texttt{is-a}\,Z$, and such embeddings under coordinatewise
dominance capture this transitivity. Other ontological relations
are also transitive, such as \texttt{part-of} and causes, so embeddings that
represent this structural property are of particular interest.

Note that the use of the transitive property of lexical entailment
is a simple species of reasoning (indeed, lexical entailment is
a special case of NLP entailment), and so exploiting transitivity
has the potential to both enable better learning, and improve reasoning.

\section{Order embedding algorithm}

\subsection{Conceptual description}

We give a construction for building an approximate embedding for terms with
a partial order relation, such as might be given in an
ontology. That is, suppose $(U,\preceq)$ is a partially ordered
set. Our approach, conceptually, is for all $x\in U$:
\begin{enumerate}
    \item \label{alg step upper}
    Construct the upper set \cite{wiki-upper} of $x$, that is,
    $\upset{x} \defeq \{y \mid x\preceq y\};$
    \item\label{alg step char}
    Construct the characteristic vector $\mvec(x)\in\{0,1\}^U$ of the upper set of $x$,
    that is,
        \[
        \mvec(x)_y \defeq
        \begin{cases} 
            1 & y\in\upset{x};\\
            0 & \mathrm{otherwise}.
        \end{cases}
        \]
    \item \label{alg step sketch}
    Construct the sketch vector $\sketch(\mvec(x))\in\Reals^d$, for a target dimension $d$,
    built so that for $x,y\in U$,
    \[
    \sketch(\mvec(x))\cdot \sketch(\mvec(y)) \approx \mvec(x) \cdot \mvec(y).
    \]
\end{enumerate}
Our motivation for considering the upper sets $\upset{x}$ is that
they represent the partial order via the subset relation, that is,
\begin{align*}
x\preceq y
       & \iff \upset{x}\supseteq \upset{y}
    \\ & \iff |\upset{x}\cap\upset{y}| = |\upset{y}|.
\end{align*}
Since the characteristic vectors $\mvec(x)$ have the property
that $\mvec(x) \cdot \mvec(y) = |\upset{x}\cap \upset{y}|$,
we have that
\begin{align*}
    x\preceq y
        \iff \mvec(x) \cdot \mvec(y) = \mvec(y) \cdot \mvec(y),
\end{align*}
so the $\mvec(x)$ give a direct representation of $(U,\preceq)$
via their dot products. However, they are vectors in $|U|$ dimensions,
which is typically too big to be useful, so in the third step
above, we apply a \emph{sketching} operation, described next,
that maps the $\mvec(x)$ to lower-dimensional vectors,
while approximately preserving dot products.

\subsection{Sketching}

There are a variety of sketching methods,
but the one we have investigated the most
is \emph{countsketch} \cite{charikar2002finding}; 
here
we need two hash functions
\begin{align*}
    h_1 :\, &U \rightarrow [d] \\
    h_2 :\, &U\rightarrow \{-1,+1\},
\end{align*}
where $[d]\defeq \{1,2,\ldots d\}$.
In the ideal setting, $h_1$ and $h_2$
are uniformly random, over $[d]$ and $\{-1,+1\}$
respectively.
Given vector~$v\in\{0,1\}^U$,
its sketch in $\Reals^d$, via countsketch,
has coordinates
\[
\sketch_c(v)_i \defeq \sum_{y:h_1(y)=i} v_y h_2(y),
\]
that is, the sum of the bit flips $h_2(y)$,
over the $y\in U$ such that $v_y=1$
and $y$ hashes (using $h_1$) to~$i$.

\subsection{More concrete description}

Short-cutting the conceptual discussion,
our embedding
can be described as follows:
given $x\in U$,
$\OS{x}$
is the vector with
\begin{equation}
    \OS{x}_i  \defeq \sum_{\substack{x\preceq y\\h_1(y)=i}} h_2(y), \mathrm{\ for\ } i\in [d].
\end{equation}
If $h_1$ and $h_2$ are indeed random,
then one can show that for $v,w\in\Reals^U$,
$\sketch_c(v) \cdot \sketch_c(w)$
is an unbiased estimator of $v\cdot w$, that is,
\[
\mathbb{E}[\sketch_c(v) \cdot \sketch_c(w)]
    = v\cdot w.
\]
In particular, if $v$ and $w$ are very sparse,
so that there are no collisions in their
sketches, that is, $h_1(y) \neq h_1(y')$
for any $y,y'$ with $v_y=v_{y'}=1$ or
$w_y=w_{y'}=1$, then 
${\sketch_c(v) \cdot \sketch_c(w) = v\cdot w}$.
More generally, the sparser $v$ and $w$ are,
the more accurate the sketch-based estimate
of the dot product will be.

The role
of the bit-flip hash function $h_2$
is to reduce the effect of collisions,
by averaging out their effects, but
with enough sparsity, $h_2$ is not needed.

Other possible sketching methods involve
simhash \cite{charikar2002similarity} and minhash \cite{broder1997resemblance}, 
for which the application to dot-product estimation
is less direct, and processing is more complex
and expensive, but which may well be more effective
in sketching vectors that are not sparse.
Our preliminary experiments suggested that simhash
is not competitive with countsketch
for the cases we considered.

\section{$\mathtt{OrderSketch}$ applied to WordNet}

We apply our order embedding technique to WordNet \cite{miller1998wordnet},
mainly to the hypernym relation. Here
the relation is between specific word meanings,
where a word meaning is represented in WordNet as a \texttt{synset}.
Our hashes $h_1$ and $h_2$ are implemented as Python's
native \texttt{hash} applied to the \texttt{synset}s and to
\texttt{synset} names, respectively. The upper sets 
$\upset{x}$ are, for each \texttt{synset}, the set of all
\texttt{synset}s that hypernyms of it (including transitively),
and each lemma name, considered as a union of \texttt{synset}s,
is represented as the union of their upper sets.

We obtain representations $\OS{x}$
of \texttt{synset}s and lemma names, using countsketch.
When \texttt{synset} $y$ is a hypernym of lemma name $x$,
$|\upset{x}\cap\upset{y}| = |\upset{y}|$,
by construction, and so we expect
\[
R_{x,y}\defeq \frac{\OS{x}\cdot \OS{y}}{\OS{y}\cdot \OS{y}}
\]
to be close to one. If $x$ and $y$
are not related, we expect $R_{x,y}\approx 0$.

To test our representation, we consider $R_{x,y}$
in these two situations, where we consider
for the positive case
all \texttt{synset}s $y$, and all lemma names $x$ in WordNet
that have one sense (\texttt{synset}) that is a hyponym of $y$.
Our proxy for the negative,
unrelated case, is choosing for each \texttt{synset} $y$
a number of lemma names $x$ at random, and
computing $R_{x,y}$.

\begin{wrapfigure}[15]{R}{0.5\textwidth}
    \centering
	\includegraphics[width=6.5cm]{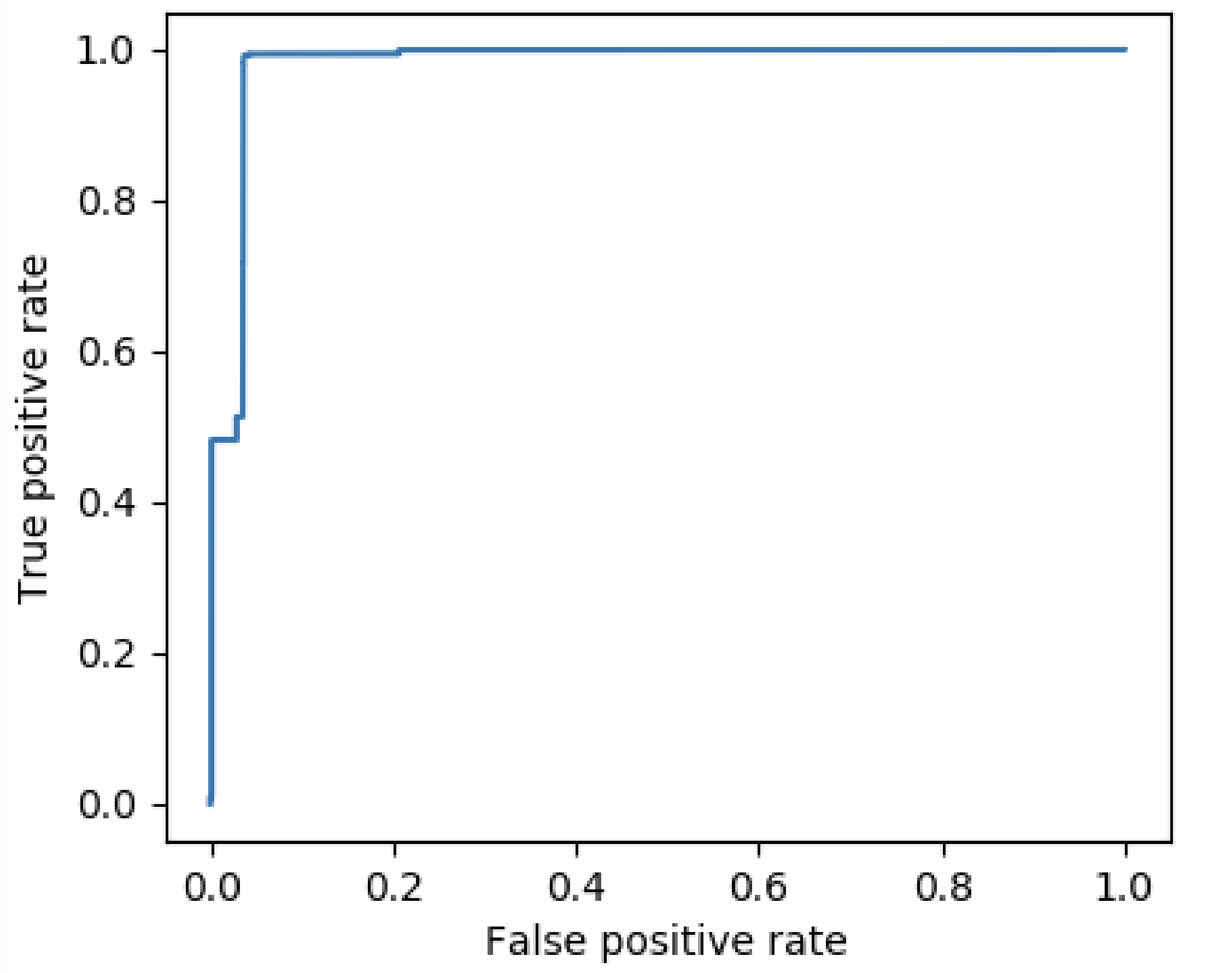}
	\caption{ROC of hypernymy classification based on $\mathtt{OrderSketch}$; the area under
	the curve is 0.9733.}
	\label{fig wn_roc}
\end{wrapfigure}

We used this scheme with embedding dimension $d=100$,
and 20 random $x$ per \texttt{synset} in the negative case.
The average deviation from $R_{x,y}=1$ in the positive
case was $6\%$, and from zero in the negative case
was $9\%$.
By considering
a decision rule $R_{x,y}\ge T$ for different $T$,
we obtain a range of true and false classifications,
with ROC curve shown in Figure~\ref{fig wn_roc}.
We also explored AUROC, the area under the ROC,
as a function of $d$; see Figure~\ref{fig wn_roc} in the Appendix,
and further results in Section~\ref{subsec wn eval} there.
A~similar experiment for \texttt{part-of},
that is, the part-meronym relation, yielded a classifier
with 0.985 AUROC, and AUROC 0.986 for
the substance-meronym relation.

\section{$\mathtt{OrderSketch}$ applied to SnoMeSHNet}
Before beginning with applying the embedding algorithm to medical ontologies just as with Wordnet,
we needed to perform an intermediate step of organizing the ontology elements as \texttt{synset}s
and lemmas, during which we formed SnoMeSH, in which we composed related elements
across SNOMEDCT and MeSH \citep{rogers1963medical}
The ordering information in SnoMeSH is based on the \texttt{is-a} relationship
in SNOMEDCT, as it aligns more with how hypernymy is represented in Wordnet, although there
are more ordering relations such as \texttt{part-of} to exploit as well. 

It can also be noted that the provision of synonymous concepts from one ontology as the lemmas of a \texttt{synset}, corresponding to matching a concept of another ontology,
enables contextual information to be captured along with ordering information.
While this is explicit during construction of the SnoMeSH dataset,
such synonymy amidst hypernymy can also be present in \texttt{synset}s of Wordnet
for our embedding algorithm to capture. One more point to notice is that unlike Wordnet,
where a single lemma can have multiple senses, here in SnoMeSH we assume there is only one medical sense, although there is potential in future to explore the granularity
of sense even within medical usage.

We obtain representations $\OS{x}$ of \texttt{synset}s and lemma names of SnoMeSHNet, using countsketch. We use 500,000 \texttt{synset}s, out of which 100,000 have more than one lemma, and produced an embedding of dimension $d=500$. The area under the ROC curve for this is 0.9723. Further results are given
in Section~\ref{subsec sno} in the Appendix.

\subsection{Additional findings from SnoMeSH}
While building an order embedding using the SnoMeSH dataset, we discovered loops in
the \texttt{is-a} relationship in SNOMEDCT, which we currently handle by detecting strongly connected components and merging them into single \texttt{synset}s,
also altering the concerned \texttt{synset}'s hypernyms and hyponyms. This also
presents an opportunity to involve a human-in-the-loop domain expert,
to verify these bad links and directly fix shortcomings in ontologies.
In addition to the ordering information provided through SNOMEDCT, ordering information can also be derived from MeSH through applied heuristics: when a MeSH concept has
synonyms, and the synonyms are not given as synonyms of each other, then the initial
MeSH concept can be regarded as the hypernym of each of these synonyms, with all the hypernymy chains being of depth 1. Due to the low depth, we preferred not to use it
for our experiments, but is still worth noting.

\section{Execution times}
The experiments were run on a personal Macbook pro with 16GB memory and processor 2.7 GHz Quad-Core Intel Core i7 with no graphics processor engaged. The following observations were made: There was an average 103.2\% increase in end-to-end execution time for a doubling up of dataset size (number of \texttt{synset}s with hypernymy relations) to be processed, while embedding dimension was kept the same. There was an average of 47.3\% increase in end-to-end execution time for a doubling up of embedding dimension, with size of \texttt{synset}s maintained. A dataset of 500,000 \texttt{synset}s with 100,000 of them having more than one lemma,
and an embedding dimension of 300, takes an end-to-end execution time of 901.898 seconds, that is, about 15 minutes. See also Section~\ref{subsec exec}.

\section{Future experiments}
We have shown that embeddings that embody ordering knowledge from ontolgies are
readily built; we plan to use these embeddings standalone or in combination with contextual embeddings on real-world applications and benchmarks such as BLESS dataset.
In addition to widely used benchmarks for this category of embedding technique,
we also plan to test the usefulness of our embedding on specialty real-world
applications with sufficient vocabulary overlap, such as on medical applications,
including disease code detection on MIMIC \textit{NOTEEVENTS},
and other textual data with sufficient vocabulary overlap with SNOMEDCT
and MeSH concepts. In addition to using our technique for feature representation,
we also plan to use our embedding algorithm to detect cycles in ontologies when the ontology is intended to be hierarchical but is not (as with SNOMEDCT for example),
infer relationships within it (as in MeSH) and across (as in SnoMeSH) ontologies.
We plan to do this more automatically, rather than involving applying heuristics as we do currently, and use these verified and established relationships to suggest completion
to properties of entities within and across ontologies.

Given the quick execution time and ability to run on any configuration of even personal systems, we hope to have our embedding algorithm deployed where domain experts, such as doctors in the case of medical ontology embedding, can add or edit their local version
of the ontology, and then retrain and obtain new embeddings for text processing
nearly instantaneously.
We would also like to develop prototype systems to drive this line of usage.

\section{Conclusion}
In this work, we presented a low-carbon-footprint method to capture ordering knowledge from ontologies of any given domain from partial orders, specifically \texttt{is-a} relationships, and amending this knowledge using synonymy for \texttt{synset}s or lemmas. Along with this, we provided some insights into what the graphical representation of
the embeddings tell us regarding our current experiments, and the scope for future work.

\bibliography{anthology,emnlp2020}
\bibliographystyle{acl_natbib}

%
%
\newif\ifSTANDALONEAPP
\STANDALONEAPPfalse

\ifSTANDALONEAPP
\documentclass[11pt,a4paper]{article}
\usepackage[hyperref]{emnlp2020}
\usepackage{times}
\usepackage{latexsym}
\renewcommand{\UrlFont}{\ttfamily\small}
\usepackage{amssymb}
\usepackage{amsmath}
\usepackage{graphicx}
\usepackage{makecell}

\usepackage{microtype}



\newcommand{\defeq}{\stackrel{\textit{\tiny{def}}}{=}}
\newcommand{\Reals}{{\mathbb R}}
\newcommand{\mvec}{\mathrm{vec}}
\newcommand{\sketch}{\mathrm{sk}}
\newcommand{\upset}[1]{{\uparrow}\{#1\}}
\newcommand{\OS}[1]{\mathtt{\small OrderSketch}(#1)}

\newcommand\BibTeX{B\textsc{ib}\TeX}

\title{Fast and Simple Order Embeddings from Ontologies - Supplementary Material}

\author{Ken Clarkson \\
  IBM Research / Address line 1 \\
  Affiliation / Address line 2 \\
  Affiliation / Address line 3 \\
  \texttt{email@domain} \\\And
  Sanjana Sahayaraj \\
  Affiliation / Address line 1 \\
  Affiliation / Address line 2 \\
  Affiliation / Address line 3 \\
  \texttt{email@domain}}
\date{}

\begin{document}
\maketitle

\fi 

\appendix

\section{Appendices}

\subsection{Creating the SnoMeSH dataset}
For each SNOMEDCT concept, a compound BioOntology \citep{bioont} query containing tokens of the concept is launched on MeSH ontology. Out of the accepted MeSH return results that contain both \emph{prefLabel} and \emph{synonym} attributes, the results are further validated such that there is an overlap of atleast one token between the snomed concept and MeSH \emph{prefLabel} and between the snomed concept and at least one \emph{synonym}. This ensures sufficient contextual match of snomed entry and MeSH entry and also between the MeSH \emph{prefLabel} and it's \emph{synonym}s with respect to the snomed concept.

\begin{wraptable}{r}{0.55\textwidth}
    \centering
    \begin{tabular}{l|l}
    \hline \textbf{MeSH synonyms} & \textbf{MeSH prefLabel} \\ \hline
    \makecell{'Cranium', 'Skulls',  \\ 'Calvaria', 'Calvarium'} & Skull \\ \hline
    \makecell{'Suture Technique', \\ 'Technique, Suture',..., \\'Technics, Suture' } & Suture Techniques \\ \hline
    \makecell{'Fractures, ... 'Non-Depressed \\ Skull Fractures','Linear \\ Skull Fractures' } & Skull Fractures \\ \hline
    \end{tabular}
    \caption{Example for how how a single SNOMEDCT concept \textbf{Entire occipitomastoid suture of skull (body structure)} gets augmented in SnoMeSH}
    \label{example-table}
\end{wraptable}

An example of some SnoMeSH entries for the SNOMEDCT concept of \textbf{Entire occipitomastoid suture of skull (body structure)} are presented in Table~\ref{example-table} to show how the SnoMeSH entries can provide contextual knowledge on related anatomical parts, medical procedures and related medical findings for this instance. Since MeSH terms are more descriptive labels than the phrase like nature of SNOMEDCT concepts, the MeSH entries that pass the validation are most likely to contain essential information and exclude returning non-informative parts like prepositions and determinants.

Also SNOMEDCT is noisier compared to MeSH wherein it has related entries such as \texttt{Sandals (physical object) \texttt{is-a} Footwear (physical object)} which is not much relevant to medical domain. The MeSH match and curation step will not include this entry in SnoMeSHNet and hence also excluded from \texttt{synset} and lemmas used by our embedding algorithm. In a way, the amending of SNOMEDCT using MeSH to create a SnoMeSH entry, is a realization for curating more domain relevant entries, since now we have two domain specific ontologies, to verify the relevance of a particular concept against.

\subsection{Evaluating our representations}

As discussed in the
main text, to test our representations, we consider $R_{x,y}$
in two situations, where we consider
for the positive case
all \texttt{synset}s $y$, and all lemma names $x$ in WordNet
that have at least
one sense (\texttt{synset}) that is a hyponym of $y$.
Our proxy for the negative,
unrelated case, is choosing for each \texttt{synset} $y$
a number of lemma names $x$ at random, and
computing $R_{x,y}$.

One aspect we will address in future work is
to ensure that when $x\preceq y$, the approximations
we use do not result in our reporting that $y\preceq x$.
A scheme to do this is as follows: rather than 
$R_{x,y}$, we store for each $x$ also $N_x\defeq |\upset{x}|$,
and test
\[
\hat R_{x,y} \defeq \OS{x}\cdot \OS{y}/N_y,
\]
instead
of using
\[
\OS{y}\cdot\OS{y}\approx N_y.
\]
Moreover, we test also if $N_y < N_x$,
and only report $x\preceq y$ if that is true,
in addition to $\hat R_{x,y}\approx 1$.
Such a scheme would be more accurate, at the
cost of a representation with one more coordinate,
and a more complicated representation.

\subsection{Evaluating WordNet representations}\label{subsec wn eval}

\begin{figure}
\begin{subfigure}[b]{0.29\textwidth}
    \centering
	\includegraphics[width=\textwidth]{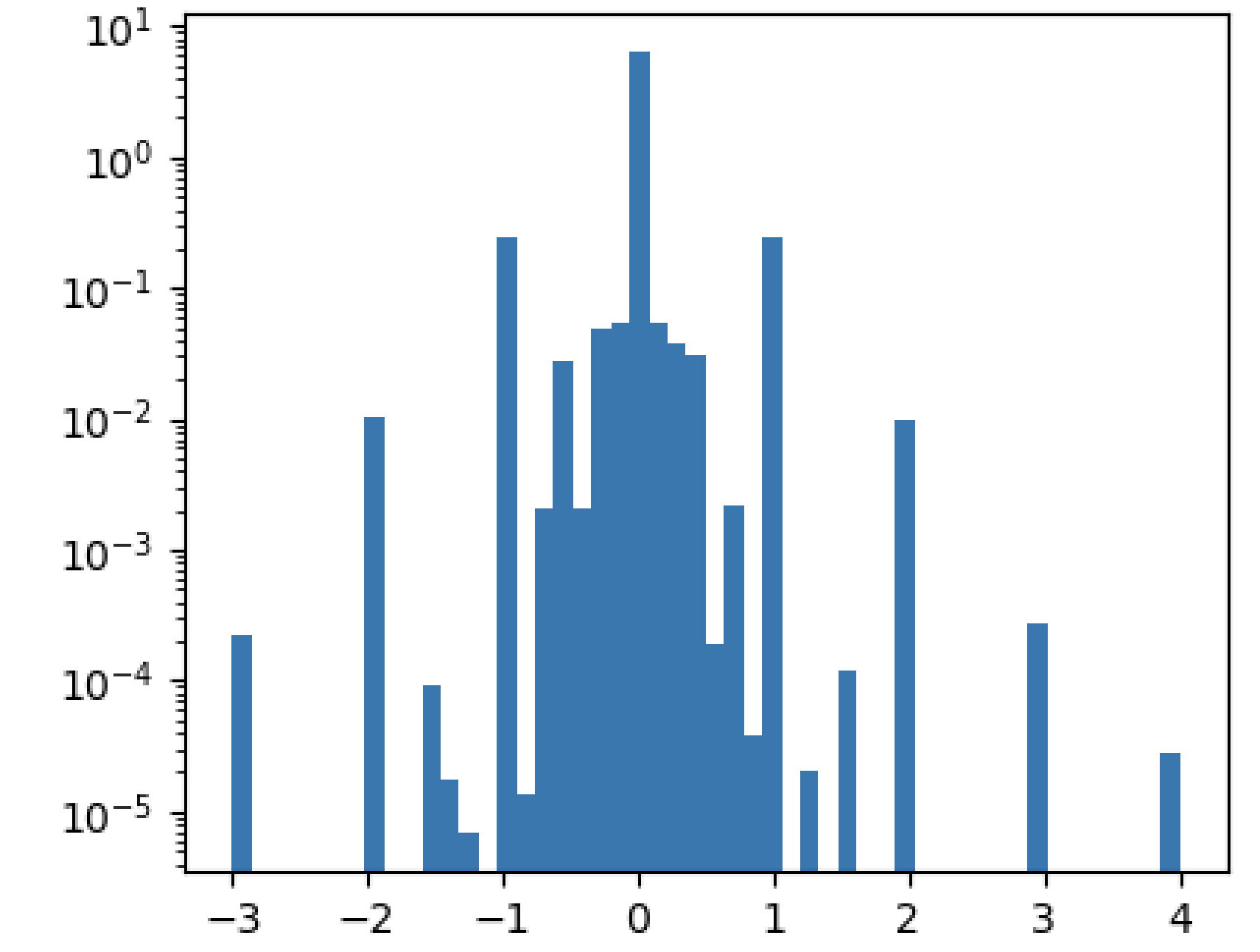}
	\caption{Distribution of $R_{x,y}$, negative case; average deviation
	from zero is 9\%.}
	\label{fig wn_neg}
\end{subfigure}
\hfill
\begin{subfigure}[b]{0.29\textwidth}
    \centering
	\includegraphics[width=\textwidth]{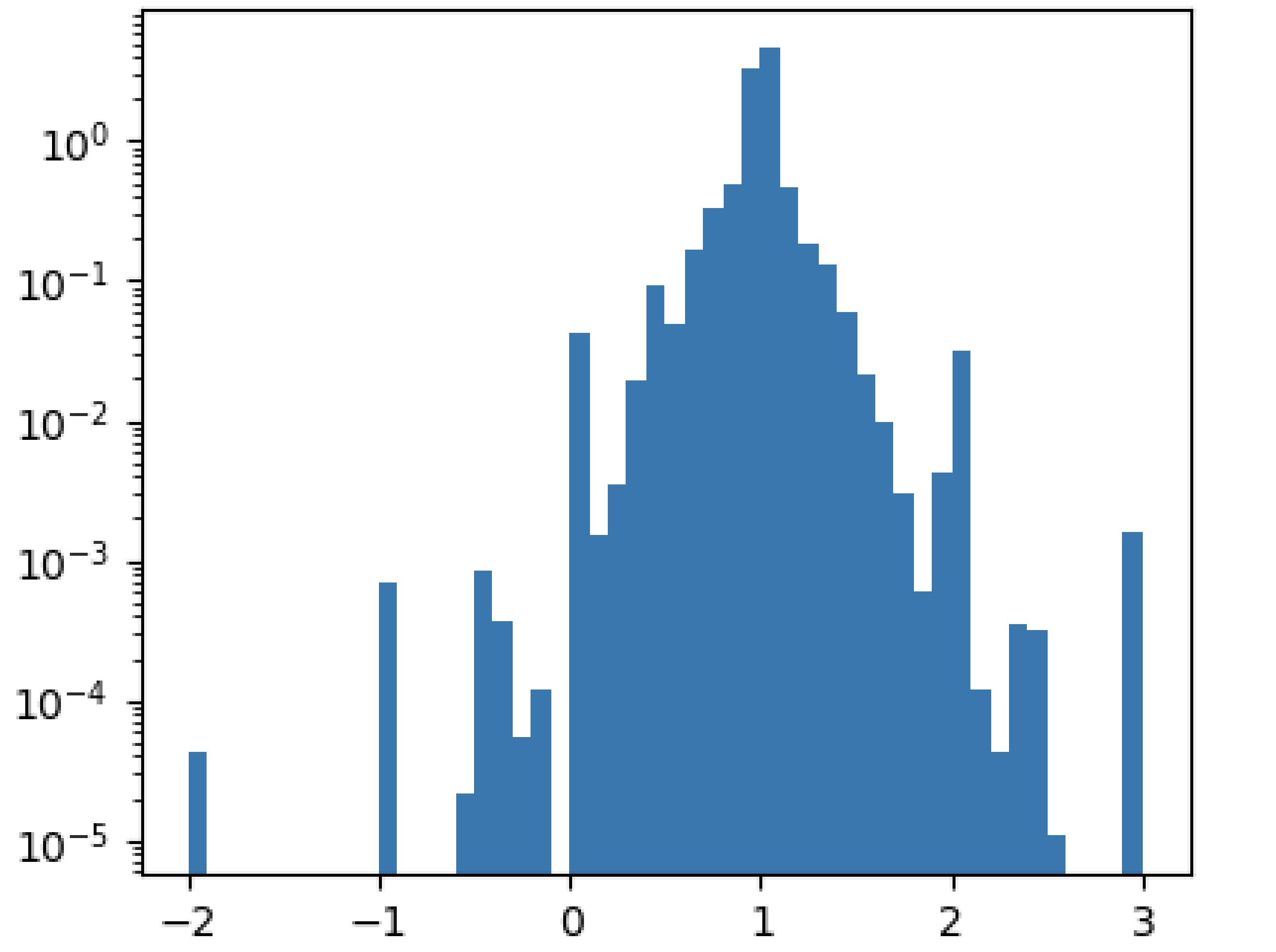}
	\caption{Distribution of $R_{x,y}$, positive case; average deviation
	from one is 6\%.}
	\label{fig wn_pos}
\end{subfigure}
\hfill
\begin{subfigure}[b]{0.29\textwidth}
    \centering
	\includegraphics[width=\textwidth]{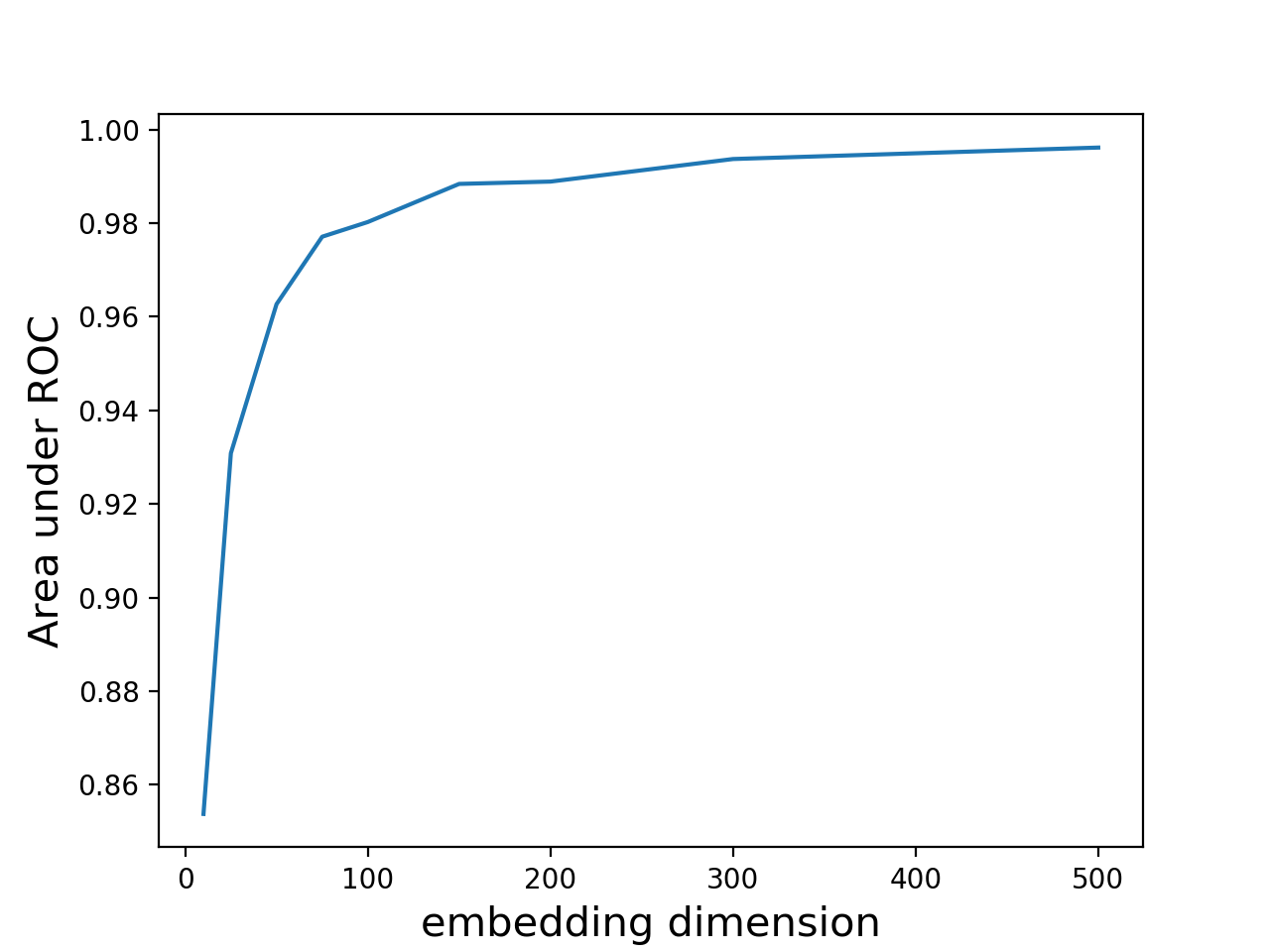}
	\caption{Area under ROC as a function of embedding dimension $d=10\ldots,500$.}
	\label{fig wn_aocs}
\end{subfigure}
\caption{Performance for Wordnet.}
\label{fig wn}
\end{figure}

For wordnet, we used our evaluation scheme with embedding dimension $d=100$,
and 20 random $y$ per \texttt{synset} in the negative case.
The average deviation from $R_{x,y}=1$ in the positive
case was $6\%$ (shown in Figure~\ref{fig wn_neg}), and from zero in the negative case
was $9\%$ (shown in Figure~\ref{fig wn_pos}).

For convenience,
in all our tests, we ignore the verbs in WordNet, as they have a loop.

As discussed in the main text, we also computed the
AUROC (area under the ROC), for part and substance holonyms
as well as hypernyms. For hypernyms, we computed
the AUROC for different values of~$d$, from 10 to 500;
please see Figure~\ref{fig wn_aocs}.

\subsection{Evaluating SnoMeSHNet representations}\label{subsec sno}
\begin{figure}
\centering
\begin{subfigure}[b]{0.29\textwidth}
    \centering
    \includegraphics[width=\textwidth]{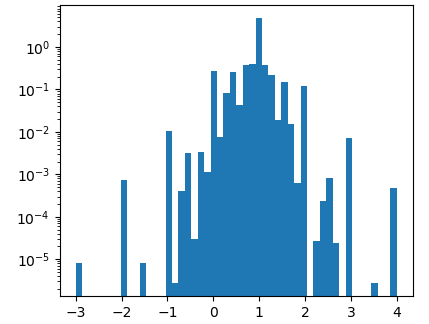}
	\caption{Distribution of $R_{x,y}$, positive case; average deviation
	from one is 15\%.}
	\label{fig snomesh_pos}
\end{subfigure}
\hfill
\begin{subfigure}[b]{0.29\textwidth}
    \centering
    \includegraphics[width=\textwidth]{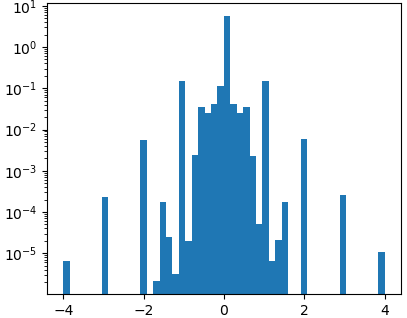}
	\caption{Distribution of $R_{x,y}$, negative case; average deviation
	from zero is 7\%.}
	\label{fig snomesh_neg}
\end{subfigure}
\hfill
\begin{subfigure}[b]{0.29\textwidth}
    \centering
    \includegraphics[width=\textwidth]{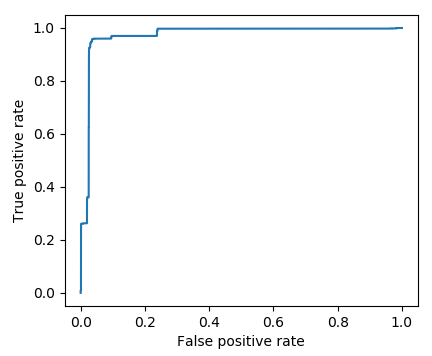}
	\caption{ROC of \texttt{is-a} classification based on $\mathtt{OrderSketch}$; the area under
	the curve is 0.9723.}
	\label{fig snoMeSH_roc}
	\end{subfigure}
	\caption{Results for SnoMeSHNet}
	\label{fig sno}
\end{figure}

For SnoMeSHNet, we used our evaluation scheme with embedding dimension $d=500$.
The average deviation from $R_{x,y}=1$ in the positive
case was $15\%$ (shown in Figure~\ref{fig snomesh_pos}), and from zero in the negative case
was $7\%$ (shown in Figure~\ref{fig snomesh_neg}).
By considering
a decision rule $R_{x,y}\ge T$ for different $T$,
we obtain a range of true and false classifications,
with ROC curve shown in Figure~\ref{fig snoMeSH_roc}.

\subsection{Execution times}\label{subsec exec}

The experiments were run on a personal Macbook pro with 16GB memory and processor 2.7 GHz Quad-Core Intel Core i7 with no graphics processor engaged. The following observations were made: 
\begin{itemize}
    \item There was an average of 103.2\% increase in end-to-end execution time for a doubling up of dataset size (number of \texttt{synset}s with hypernymy relations) to be processed while embedding dimension was kept the same.
    \item There was an average of 6.4\% increase in end-to-end execution time for a 10-fold increase in the number of \texttt{synset}s that had more than one lemma, while size of \texttt{synset}s and embedding dimension were maintained.
    \item There was an average of 47.3\% increase in end-to-end execution time for a doubling up of embedding dimension, with size of \texttt{synset}s maintained.
    \item A dataset of 500,000 \texttt{synset}s with 100,000 of them having more than one lemma, and an embedding dimension of 300, takes an end-to-end execution time of 901.898 seconds, that is, about 15 minutes.
\end{itemize}
A few profiling entries from an array of experiments we ran with SnoMeSHNet can be found in Table~\ref{times-table}.
\begin{table}[t]
    \centering\small
    \begin{tabular}{l|l|l}
        \hline \textbf{No. of \texttt{synset}s} & \textbf{Component} & \textbf{Secs} \\ \hline
        10,000 & Hypernym chains & 0.038 \\
        100,000 & Hypernym chains & 0.687 \\
        500,000 & Hypernym chains & 4.714 \\ \hline
        10,000 & No loops seen & 0.000 \\
        100,000 & Fix 27 loops & 0.006 \\  
        500,000 & Fix 835 loops & 0.049 \\ \hline
        10,000 & Two hash functions & 0.062 \\
        100,000 & Two hash functions & 0.674 \\
        500,000 & Two hash functions & 13.298 \\ \hline
        10,000 & Create vectors & 0.455 \\
        100,000 & Create vectors & 4.534 \\
        500,000 & Create vectors & 55.988 \\ \hline
        10,000 & End-to-end $d=100$ & 16.539 \\
        100,000 & End-to-end $d=200$ & 207.712 \\
        500,000 & End-to-end $d=300$ & 901.898 \\ \hline
    \end{tabular}
    \caption{Profiling across different size runs, components and embedding dimensions}
    \label{times-table}
\end{table}

\ifSTANDALONEAPP

\bibliography{anthology,emnlp2020}
\bibliographystyle{acl_natbib}
\end{document}

\fi 

\end{document}